\documentclass[runningheads]{llncs}
\usepackage[T1]{fontenc}
\usepackage{graphicx,verbatim}
\usepackage{amsmath,amssymb}
\usepackage{booktabs}
\usepackage{multirow}
\usepackage{marvosym}
\usepackage{hyperref}
\newcommand{\best}[1]{\textbf{#1}}
\newcommand{\second}[1]{\underline{#1}}
\begin{document}
\title{Counterfactual Anatomy-guided Spatial-Temporal Decoding for Annotation-Free Hallucination Mitigation in Medical VLMs}
\titlerunning{Counterfactual Anatomy-guided Spatial-Temporal Decoding}
%
\author{Yifan Lu \and
Adinath Dukre \and
Abhijit Das \and
Ziyun Zou \and
Haolin Yang \and
Yutong Xie \and
Imran Razzak\textsuperscript{\Letter}}
\authorrunning{Lu et al.}
%
\institute{Mohamed bin Zayed University of Artificial Intelligence, Abu Dhabi, UAE \\
 MedOS, Abu Dhabi, UAE \\
\email{Imran.Razzak@mbzuai.ac.ae}}
\maketitle              
\begin{abstract}
Medical vision-language models (Med-VLMs) have demonstrated strong performance on medical visual question answering, yet they remain prone to hallucination, generating clinically unsupported statements that are insufficiently grounded in image evidence. Mitigation methods applied during decoding offer a practical solution, but they typically lack anatomical awareness or rely heavily on ground truth annotations, which limits their applicability. We propose Counterfactual Anatomy-guided Spatial-Temporal decoding (CAST), a framework that operates entirely during inference and requires no manual annotations for anatomically grounded hallucination mitigation. CAST automatically discovers anatomical regions relevant to the given query through broad medical segmentation. It then selects a compact, causally informative area using counterfactual intervention based on the drop in answer likelihood under occlusion. Guided by this chosen region, CAST performs a unified contrastive decoding process, combining classifier-free guidance to correct spatial attention with stepwise temporal contrast to regulate generation dynamics. Experiments on the SLAKE and MIMIC-CXR datasets across three Med-VLMs demonstrate that CAST consistently outperforms strong baselines and surpasses decoding strategies reliant on ground truth. Our results indicate that compact, automatically selected regions provide highly effective contrastive guidance without expert annotations, offering a practical and generalizable solution for improving spatial grounding and reducing hallucinations. Code is available at \href{https://github.com/csyifan/CAST}{https://github.com/csyifan/CAST}.

\keywords{Medical Vision-Language Models \and Hallucination Mitigation \and Contrastive Decoding \and Causal Discovery}
\end{abstract}

\section{Introduction}
Medical vision-language models (Med-VLMs) have recently demonstrated impressive performance on medical visual question answering (Med-VQA) by jointly reasoning over images and clinical language~\cite{medmo,flm,hulu}. Despite their strong average accuracy, these models remain prone to hallucination, generating fluent but clinically unsupported statements that lack sufficient grounding in the image~\cite{dimmm,dina,svv}. In medical settings, such errors can have serious consequences, as a single hallucinated finding may mislead clinical interpretation or subsequent decision-making~\cite{utt,bibmm,mk2}. Improving faithfulness without expensive retraining has therefore become a central challenge for the reliable deployment of Med-VLMs~\cite{mmc,allin,time}.

Decoding-time methods are attractive because they are plug-and-play and do not require additional supervision~\cite{icll,ttr,npp}. Visual Contrastive Decoding (VCD)~\cite{leng2024vcd} mitigates hallucinations by contrasting predictions from the original image and a noise-perturbed version, encouraging tokens that remain stable under visual corruption. Decoding by Contrasting Layers (DoLA)~\cite{dola} improves factuality by contrasting logits from early and late transformer layers, suppressing tokens driven by memorized knowledge rather than contextual evidence. OPERA~\cite{opera} penalizes over-confident attention patterns and reallocates attention during generation to reduce over-trust in spurious visual cues. Although effective for general-purpose VLMs~\cite{kang2026vispec}, these methods are largely anatomy-agnostic and apply global corrections that do not explicitly account for the localized nature of medical evidence. Anatomical Region-Guided Contrastive Decoding (ARCD)~\cite{arcd} addresses this limitation by introducing spatial grounding through classifier-free guidance (CFG) conditioned on an anatomical region-of-interest (ROI). At each decoding step, ARCD contrasts a full-image branch with an ROI-suppressed branch, steering generation toward tokens supported by the diagnostic region. However, ARCD depends on ground-truth (GT) ROI annotations, which are costly, scarce, and dataset-specific, limiting real-world applicability.

We investigate whether region-guided contrastive decoding can be made annotation-free without sacrificing effectiveness by automatically discovering question-relevant ROIs for guidance. Empirically, we find that such automatically selected regions not only remove the dependency on expert annotations but can also provide stronger guidance in practice. This is partly because the effectiveness of ROI-guided CFG depends on mask granularity: ground-truth bounding boxes in Med-VQA benchmarks are often coarse and high-coverage, and suppressing them removes most visual information from the unconditional branch, weakening the region-specific contrast. In contrast, compact and precise masks preserve contextual cues while selectively removing relevant evidence, yielding a cleaner contrastive signal during decoding.

Motivated by this observation, we propose \textbf{C}ounterfactual \textbf{A}natomy-guided \textbf{S}patial-\textbf{T}emporal (\textbf{CAST}) decoding, an annotation-free framework for hallucination mitigation in Med-VLMs. CAST consists of three stages. First, in the proposal stage, we generate diverse anatomical region candidates using a medical segmentation model through efficient multi-concept inference. Second, in the selection stage, we identify the most question-relevant and CFG-compatible region via a counterfactual intervention criterion that measures the likelihood drop under regional occlusion while discouraging overly large masks. Third, in the decoding stage, we introduce a spatial-temporal contrastive strategy that corrects where the model attends via anatomy-guided CFG and regulates what it predicts via step-wise temporal contrast. The entire pipeline operates at inference time and remains fully plug-and-play. Across three Med-VLMs on SLAKE and MIMIC-CXR, CAST outperforms strong decoding-time baselines and GT-dependent region-guided methods, achieving effective spatial grounding without expert annotations. 

\begin{figure}[t]
  \centering
  \includegraphics[width=\textwidth]{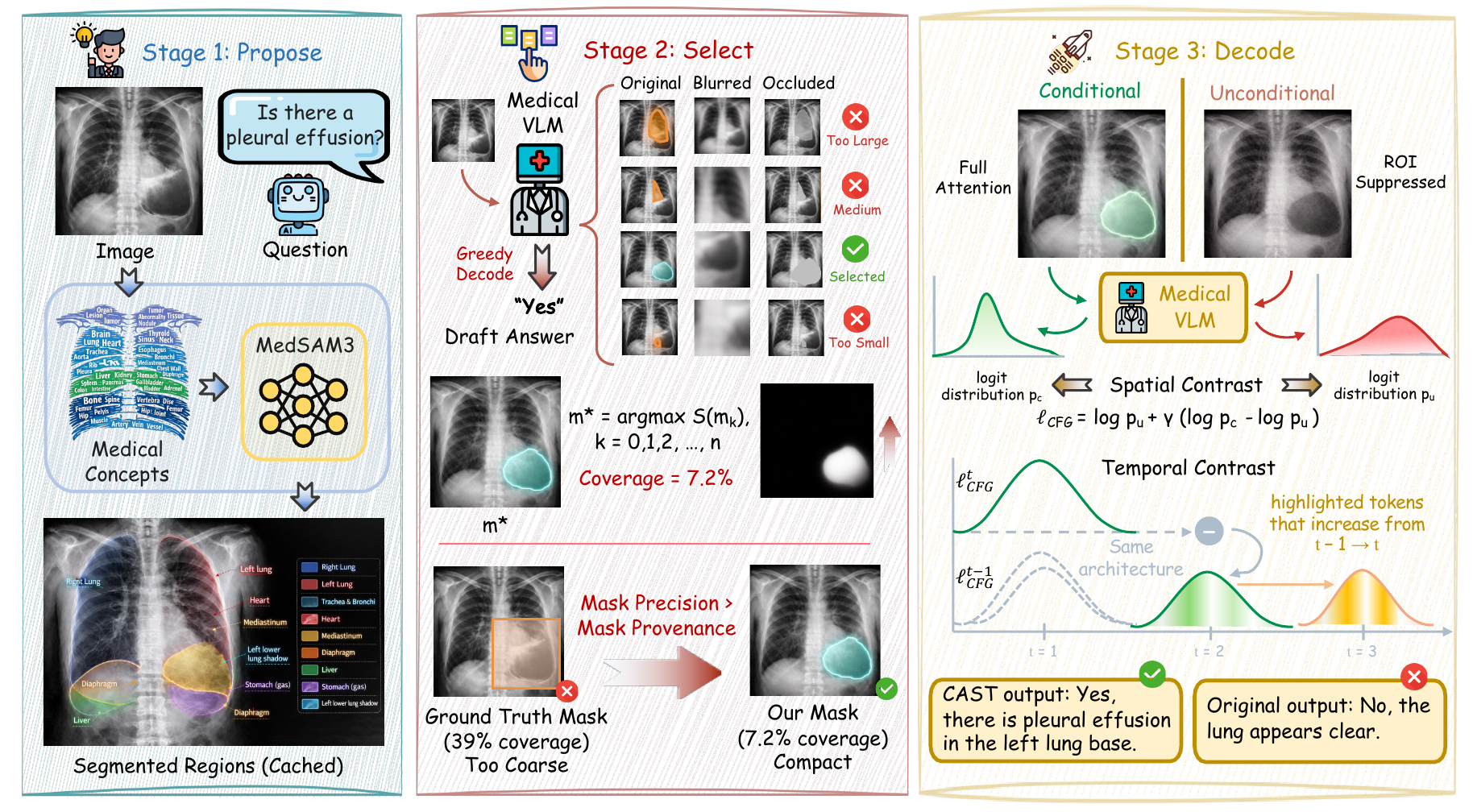}
\caption{Overview of the proposed CAST framework. It (1) proposes anatomical regions via multi-concept medical segmentation, (2) selects a compact question-relevant ROI using counterfactual likelihood drop under occlusion, and (3) applies spatial-temporal contrastive decoding: anatomy-guided CFG for grounding and step-wise temporal contrast to suppress hallucinations.}
\label{fig:pipeline}
\end{figure}

\section{Method}
CAST is an annotation-free, inference-time framework for mitigating hallucinations in Med-VLMs via automatic ROI discovery and spatial-temporal contrastive decoding. As illustrated in Fig.~\ref{fig:pipeline}, we first generate candidate anatomical masks with multi-concept medical segmentation and filter them for CFG compatibility. Given a question, we select the ROI that most affects the model's likelihood under regional occlusion while favoring compactness. Finally, CAST performs anatomy-guided CFG for region grounding and a step-wise temporal contrast to damp language-driven momentum. The entire pipeline is fully plug-and-play, requiring no retraining or expert annotations.

\subsection{Zero-Shot Anatomical Region Proposal}
\label{sec:proposal}
The first stage of CAST aims to extract a diverse set of anatomically coherent region candidates directly from the input image $\mathbf{x}$. To achieve this without dataset-specific annotations, we leverage the robust medical priors embedded in a text-guided medical segmentation model, MedSAM3~\cite{medsam3}. Rather than sequentially querying individual anatomical structures, we employ an efficient multi-concept inference strategy. We construct a comprehensive vocabulary of anatomical concepts (spanning thoracic, abdominal, musculoskeletal, and vascular systems) and encode them into a unified prompt. The model's DETR-style~\cite{detr} decoder processes this joint embedding to activate diverse parallel queries, generating segmentation masks for multiple structures in a single forward pass. To ensure the proposals are suitable for localized decoding, the raw masks undergo a compactness-aware filtering pipeline. Masks are filtered by area coverage to remove pixel-level noise and excessively global masks. We then apply spatial non-maximum suppression based on intersection over union to eliminate near-duplicates~\cite{skz}, followed by morphological filtering to discard degenerate shapes. This results in a spatially diverse and anatomically valid set of $K$ candidate regions $[m_1, \ldots, m_K]$, extracted with minimal computational overhead.

\subsection{Counterfactual Region Selection}
\label{sec:selection}
Given the $K$ candidates, CAST must autonomously identify the single region $m^*$ that is most causally responsible for the model's clinical reasoning regarding the current query $q$. We frame this as a causal discovery problem, evaluating each region via counterfactual intervention. We first generate a draft sequence $\hat{y} = (y_1, \ldots, y_T)$ utilizing greedy decoding on the factual, unoccluded image $\mathbf{x}$. To isolate the causal contribution of a candidate region $m_k$, we define a counterfactual intervention $\text{do}(\text{occlude } m_k)$ by selectively corrupting the pixel space strictly within the mask boundaries. To prevent the model from over-indexing on occlusion artifacts, we utilize two complementary operators, $o \in \{\text{blur}, \text{mean}\}$, which independently disrupt fine-grained texture and local contrast statistics. The causal effect (CE) of region $m_k$ is quantified as the drop in the predictive log-likelihood of the draft answer when the region is removed:
\begin{equation}
    \text{CE}_o(m_k) = \log p(\hat{y} \mid \mathbf{x}, q) - \log p(\hat{y} \mid \mathbf{x}_{\text{occ}}^{(o,k)}, q),
\end{equation}
where $\mathbf{x}_{\text{occ}}^{(o,k)}$ represents the image under occlusion mode $o$. Effective contrastive guidance requires spatial precision; a region that spans the entire image provides no localized contrastive signal. Therefore, our selection objective explicitly balances causal relevance with mask compactness and cross-modal consistency:
\begin{equation}
    S(m_k) = \frac{1}{2}\sum_{o} \text{CE}_o(m_k) - \lambda \cdot \text{area}(m_k) - \mu \cdot \mathrm{Var}\bigl[\text{CE}_\text{blur}(m_k), \text{CE}_\text{mean}(m_k)\bigr].
\label{eq:score}
\end{equation}
Here, the area penalty $\lambda$ enforces mask compactness for localized contrast, while $\mu$ penalizes variance to filter out occlusion-specific artifacts. The highest-scoring mask $m^* = \arg\max_k S(m_k)$ is then selected as the optimal diagnostic anchor.

\subsection{Unified Spatial-Temporal Contrastive Decoding}
\label{sec:decode}
With the optimal region $m^*$ identified, CAST introduces a lightweight, sequential decoding strategy to simultaneously correct two orthogonal hallucination axes: spatial grounding (\emph{where} the model attends) and temporal generation (\emph{what} it predicts). At each decoding step $t$, a dual-branch forward pass is constructed. The unconditional branch heavily suppresses attention to visual tokens within $m^*$, forcing reliance on surrounding context and language priors, while the conditional branch processes the full image natively. Spatially guided logits are computed as:
\begin{equation}
    \boldsymbol{\ell}_\text{CFG}^{(t)} = \log p_u^{(t)} + \gamma \cdot \bigl(\log p_c^{(t)} - \log p_u^{(t)}\bigr), 
\end{equation}
where $\gamma$ is the guidance scale, and $p_c$ and $p_u$ are conditional and unconditional token probabilities. This amplifies tokens dependent on the targeted anatomical region, reducing reliance on spurious background correlations. Following spatial grounding, CAST applies an orthogonal temporal correction across consecutive steps:
\begin{equation}
    \boldsymbol{\ell}_\text{final}^{(t)} = w_m \cdot \boldsymbol{\ell}_\text{CFG}^{(t)} - w_p \cdot \boldsymbol{\ell}_\text{CFG}^{(t-1)},
\end{equation}
where $w_m$ and $w_p$ are weighting hyper-parameters. The theoretical foundation for this temporal contrast is that the prior step's distribution, $\boldsymbol{\ell}_\text{CFG}^{(t-1)}$, heavily encodes the default trajectory driven by the model's internal language momentum. By subtracting this prior distribution from the current active prediction $\boldsymbol{\ell}_\text{CFG}^{(t)}$, CAST selectively amplifies tokens that are newly activated by the current visual evidence, while suppressing tokens that are merely passively inherited from previous steps. Because the spatial contrast modulates the source of visual attention and the temporal contrast regulates habitual token transitions, the two mechanisms operate harmoniously without interference.


\section{Experiments}

\subsection{Experimental Setup}

\textbf{Datasets and Baselines.}
We evaluate on SLAKE~\cite{slake} and MIMIC-CXR~\cite{mimic}. SLAKE contains 1,061 multi-modal (CT/MRI/X-ray) QA pairs with open- and closed-ended questions and bounding boxes. MIMIC-CXR includes 2,134 chest X-ray QA pairs, primarily closed-ended with pathology annotations. Following prior work~\cite{arcd}, we report Accuracy (closed), Recall (open), and weighted Overall performance. CAST is compared against the standard decoding baseline and four decoding-time methods: VCD~\cite{leng2024vcd}, DoLA~\cite{dola}, OPERA~\cite{opera}, and ARCD~\cite{arcd}.

\textbf{Implementation Details.}
To demonstrate the plug-and-play versatility of CAST, we apply it to three Med-VLMs: Phi-3.5V-Med~\cite{arcd}, HuatuoGPT-Vision-7B~\cite{huatuo}, and LLaVA-Med-7B~\cite{lmed}. MedSAM3 processes images at a resolution of $1008{\times}1008$. We apply NMS with an IoU threshold of $0.9$ to ensure spatial diversity, retaining up to $K{=}16$ candidate regions per question for counterfactual scoring. For the selection objective, both the area penalty factor $\lambda$ and the consistency coefficient $\mu$ are set to $0.1$. In the decoding stage, the spatial CFG employs a guidance scale of $\gamma{=}1.5$. The attention modulation weights for regions inside and outside the ROI ($\alpha$ and $\beta$, respectively) are empirically determined via grid search over the set $\{0.01, 0.1, 0.5, 1.0, 2.0, 3.0\}$. For the step-wise temporal contrast, we set the active momentum weight $w_m{=}1.0$ and the prior-step penalty $w_p{=}0.5$. All experiments are executed on a single NVIDIA A100 40GB GPU.

\begin{table}[t]
\caption{Comprehensive comparison on SLAKE and MIMIC benchmarks. Best in \textbf{bold}, second \underline{underlined} within each model block.}\label{tab:multimethod}
\centering
\setlength{\tabcolsep}{3pt}
\fontsize{7}{8.5}\selectfont
\begin{tabular}{l ccc ccc}
\toprule
 & \multicolumn{3}{c}{\textbf{SLAKE}} & \multicolumn{3}{c}{\textbf{MIMIC}} \\
\cmidrule(lr){2-4} \cmidrule(lr){5-7}
Method & Open & Closed & Overall & Open & Closed & Overall \\
\midrule
Phi-3.5V-Med & 38.58 & 61.42 & 50.00 & 14.29 & 58.12 & 47.24 \\
+ VCD & 40.16 & 61.42 & 50.79 & 15.87 & 59.16 & 48.43 \\
+ DoLA & \best{47.24} & 59.84 & \second{53.54} & 19.05 & 56.54 & 47.24 \\
+ OPERA & 36.22 & 65.35 & 50.79 & 14.29 & \second{61.78} & 50.00 \\
+ ARCD & \second{44.54} & \second{65.87} & 52.90 & \second{28.85} & 60.44 & \second{52.61} \\
 + \textbf{CAST (Ours)} & \second{44.90} & \best{81.49} & \best{59.25} & \best{33.03} & \best{67.85} & \best{59.22} \\
\midrule
HuatuoGPT-V-7B & 47.94 & 75.48 & 58.74 & 29.84 & 62.68 & 54.54 \\
+ VCD & \best{49.73} & 75.48 & \second{59.83} & 31.05 & 63.61 & 55.54 \\
+ DoLA & 43.97 & \second{78.61} & 57.55 & 29.64 & \second{65.30} & 56.46 \\
+ OPERA & 46.32 & 74.04 & 57.19 & \best{34.74} & 63.99 & \second{56.74} \\
+ ARCD & \second{48.33} & 75.00 & 58.79 & 30.97 & 63.12 & 55.15 \\
 + \textbf{CAST (Ours)} & 47.56 & \best{81.49} & \best{60.86} & \second{31.31} & \best{67.29} & \best{58.37} \\
\midrule
LLaVA-Med-7B & 39.62 & 62.50 & 48.59 & 22.52 & 53.71 & 45.98 \\
+ VCD & \second{40.00} & 62.98 & 49.01 & 22.77 & 53.71 & 46.04 \\
+ DoLA & 32.75 & \second{66.35} & 45.92 & 21.34 & \second{56.14} & \second{47.51} \\
+ OPERA & \best{40.59} & 65.38 & \second{50.31} & \second{26.17} & 54.52 & 47.49 \\
+ ARCD & 37.95 & 62.02 & 47.39 & 22.88 & 53.77 & 46.11 \\
+ \textbf{CAST (Ours)} & 36.49 & \best{73.32} & \best{50.93} & \best{27.16} & \best{59.69} & \best{51.63} \\
\bottomrule
\end{tabular}
\end{table}

\subsection{Experimental Results and Analysis}
\textbf{Comparison with Decoding-Time Methods.}
Table~\ref{tab:multimethod} reports the comprehensive comparison across three Med-VLMs on SLAKE and MIMIC-CXR. CAST consistently improves overall performance over strong decoding-time baselines (VCD, DoLA, OPERA) and GT-dependent region-guided decoding across all models. Most notably, CAST demonstrates substantial and consistent improvements in Closed-ended question accuracy, which directly reflects a model's ability to remain factually grounded and resist hallucination. For instance, when applied to Phi-3.5V-Med on SLAKE, CAST elevates closed-ended accuracy from 61.42 to a striking 81.49. On MIMIC, CAST improves both open and closed metrics simultaneously, indicating that spatial-temporal contrast benefits both factual grounding and short-form diagnostic responses. While anatomy-agnostic baselines such as DoLA, OPERA, and VCD occasionally yield isolated gains on Open-ended queries for specific models, they frequently exhibit volatile performance, trading closed-ended accuracy for open-ended recall. In contrast, CAST maintains highly competitive open-ended performance while decisively dominating closed-ended factual reasoning. Notably, CAST consistently outperforms ARCD, this empirical superiority strongly validates the core theoretical hypothesis introduced in our methodology: automatically discovered, dynamically selected, and highly compact region masks provide a cleaner, more targeted contrastive signal than coarse GT annotations. By explicitly optimizing for causal relevance and spatial compactness, CAST achieves superior spatial grounding and hallucination mitigation entirely free of expert annotation dependency.

\begin{figure}[t]
  \centering
  \includegraphics[width=\textwidth]{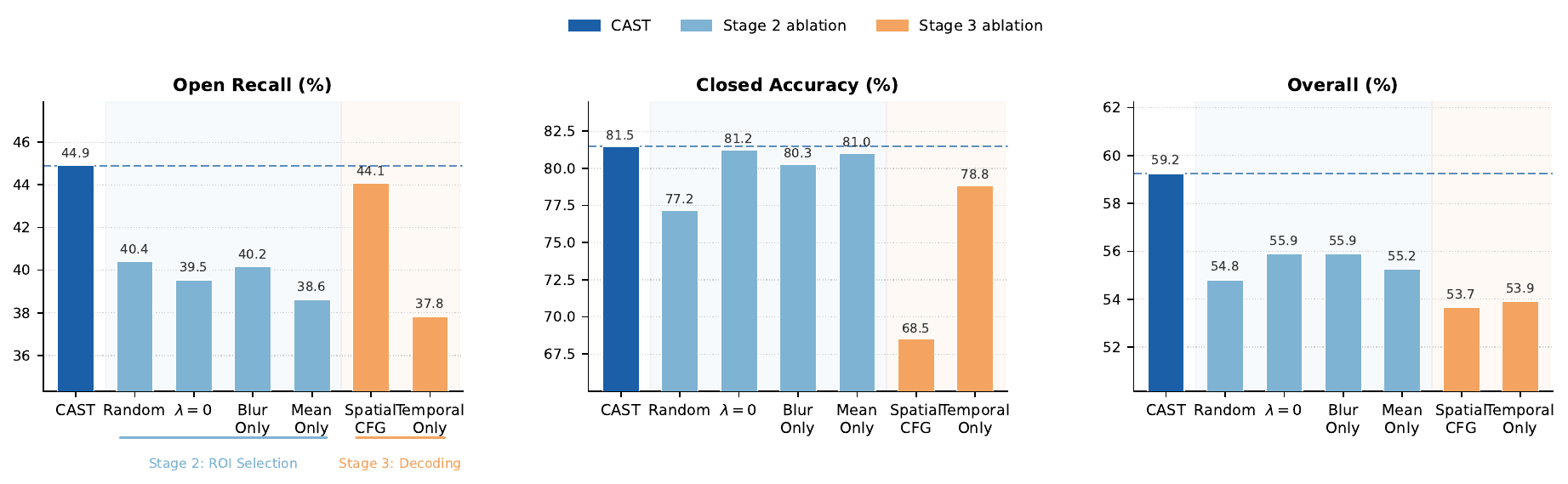}
\caption{Ablation study on SLAKE (Phi-3.5V-Med).}
\label{tab:ablation}
\end{figure}

\textbf{Ablation Study.}
Fig.~\ref{tab:ablation} ablates the core components of CAST on SLAKE, demonstrating that both our unified decoding mechanisms and counterfactual selection criteria are essential. Isolating spatial CFG or temporal contrast drops overall accuracy from 59.25\% to 53.66\% and 53.91\%, respectively, confirming their orthogonal roles: spatial CFG anchors where the model attends to rescue closed-question accuracy (from 68.51\% to 81.49\%), while temporal contrast regulates what it predicts to sustain open-question recall (44.90\% vs.\ 37.83\%). Beyond decoding, counterfactual region selection provides a critical signal; replacing it with random mask selection degrades overall accuracy by 4.4\%. Within the scoring function itself, removing the area penalty ($\lambda{=}0$) causes a severe 3.4\% drop, cementing that mask compactness is strictly required for effective CFG. Finally, relying on a single occlusion mode (blur or mean) lowers accuracy by 3.3\% and 4.0\%, highlighting the necessity of combining dual occlusions to average out modality-specific artifacts.

\begin{figure}[t]
  \centering
  \includegraphics[width=0.9\textwidth]{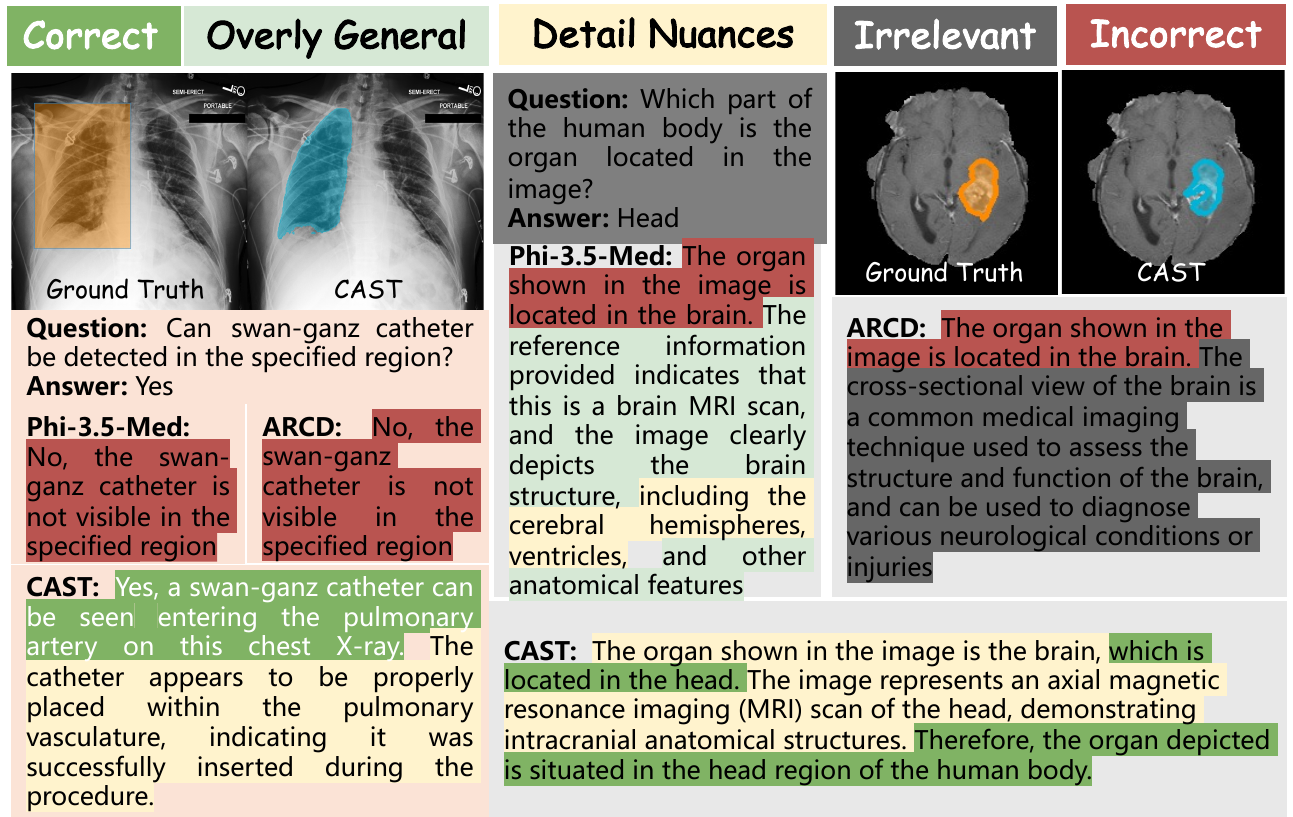}
\caption{Qualitative comparison of model-generated responses.}
\label{fig:qualitative}
\end{figure}

\textbf{Qualitative Analysis.}
Fig.~\ref{fig:qualitative} shows two failure cases where CAST overcomes limitations of coarse GT masks. In Case 1 (MIMIC X-ray), GT-ARCD fails to detect a subtle Swan-Ganz catheter because its bounding box spans the entire right lung, diluting the contrastive signal. CAST instead isolates a compact 10.6\% region around the catheter tip and correctly confirms its presence. In Case 2 (SLAKE MRI), the model must identify the anatomical region (head) rather than the highly salient organ (brain). Both the Baseline and GT-ARCD incorrectly output ``brain'' because the broad GT segmentation reinforces organ identity over spatial location, whereas CAST selects a highly compact mask highlighting specific craniofacial landmarks to correctly guide the response. Together, these cases concretely demonstrate that dynamically selected, compact masks provide superior contrastive guidance compared to static GT annotations, which can actively mislead or dilute critical visual evidence.

\textbf{Mask Compactness Analysis.}
To understand CAST’s advantage over GT-based methods, we analyze mask coverage and performance on HuatuoGPT-V-7B (SLAKE). GT bounding boxes are substantially larger than CAST-selected masks (39.41\% vs.\ 7.05\% mean coverage), with some spanning the entire image. When GT coverage exceeds 40\%, closed-question accuracy drops noticeably. Per-sample analysis shows that CAST mainly improves cases with large GT masks, while GT performs competitively only when annotations are compact. Notably, GT slightly reduces closed accuracy compared to NULL decoding, whereas CAST significantly improves it. These results indicate that oversized masks dilute the localized contrast required by classifier-free guidance, while compact, causally selected regions provide sharper and more effective signals.

\textbf{Computational Efficiency.}
As a plug-and-play inference-time framework, CAST introduces minimal overhead compared to standard decoding method. In Stage 1, MedSAM3 proposals are precomputed and cached, effectively eliminating per-question segmentation costs; even in an online setting, this overhead remains marginal relative to VLM decoding. In Stage 2, counterfactual ROI scoring operates via a lightweight teacher-forced evaluation over a short sequence ($T_{score}=32$ tokens), which is roughly an order of magnitude cheaper than a full autoregressive generation pass. Although Stage 3 employs dual-branch spatial-temporal contrast, the overall wall-clock latency of CAST remains nearly identical to that of DoLA, placing it squarely within the practical operating range of accepted inference-time methods. Furthermore, the $K$ candidate regions are independent and can be seamlessly parallelized at the engineering level for further efficiency gains.

\section{Conclusion}
In this paper, we presented CAST, a plug-and-play decoding framework for mitigating hallucinations in Med-VLMs. By formulating spatial grounding as a causal discovery problem, CAST automatically identifies compact and question-relevant anatomical regions through counterfactual intervention. We show that these dynamically selected regions address the compactness limitations of coarse ground-truth annotations and provide a sharper and more informative contrastive signal. Combined with a unified spatial-temporal decoding strategy that jointly refines visual attention and step-wise generation dynamics, CAST consistently achieves state-of-the-art performance across multiple Med-VLMs on SLAKE and MIMIC-CXR. Overall, CAST offers a practical inference-time approach to improving factual reliability without requiring expert annotations or model retraining.



\bibliographystyle{splncs04}
\bibliography{Paper}

@article{medmo,
  title={MedMO: Grounding and Understanding Multimodal Large Language Model for Medical Images},
  author={Deria, Ankan and Kumar, Komal and Dukre, Adinath Madhavrao and Segal, Eran and Khan, Salman and Razzak, Imran},
  journal={arXiv preprint arXiv:2602.06965},
  year={2026}
}

@article{flm,
  title={Fleming-VL: Towards Universal Medical Visual Reasoning with Multimodal LLMs},
  author={Shu, Yan and Liu, Chi and Chen, Robin and Li, Derek and Dai, Bryan},
  journal={arXiv preprint arXiv:2511.00916},
  year={2025}
}

@article{hulu,
  title={Hulu-med: A transparent generalist model towards holistic medical vision-language understanding},
  author={Jiang, Songtao and Wang, Yuan and Song, Sibo and Hu, Tianxiang and Zhou, Chenyi and Pu, Bin and Zhang, Yan and Yang, Zhibo and Feng, Yang and Zhou, Joey Tianyi and others},
  journal={arXiv preprint arXiv:2510.08668},
  year={2025}
}

@inproceedings{arcd,
  title={Anatomical Region-Guided Contrastive Decoding: A Plug-and-Play Strategy for Mitigating Hallucinations in Medical VLMs},
  author={Liang, Xiao and Liu, Chenxi and Ma, Zhi and Wang, Di and Jing, Bin and Wang, Quan and Shi, Yuanyuan},
  booktitle={Proceedings of the AAAI Conference on Artificial Intelligence},
  volume={40},
  number={9},
  pages={6871--6879},
  year={2026}
}

@inproceedings{dina,
  title={Vision-amplified semantic entropy for hallucination detection in medical visual question answering},
  author={Liao, Zehui and Hu, Shishuai and Zou, Ke and Fu, Huazhu and Zhen, Liangli and Xia, Yong},
  booktitle={International Conference on Medical Image Computing and Computer-Assisted Intervention},
  pages={669--679},
  year={2025},
  organization={Springer}
}

@inproceedings{leng2024vcd,
  title={Mitigating object hallucinations in large vision-language models through visual contrastive decoding},
  author={Leng, Sicong and Zhang, Hang and Chen, Guanzheng and Li, Xin and Lu, Shijian and Miao, Chunyan and Bing, Lidong},
  booktitle={Proceedings of the IEEE/CVF Conference on Computer Vision and Pattern Recognition},
  pages={13872--13882},
  year={2024}
}

@inproceedings{dola,
  title={Dola: Decoding by contrasting layers improves factuality in large language models},
  author={Chuang, Yung-Sung and Xie, Yujia and Luo, Hongyin and Kim, Yoon and Glass, James R and He, Pengcheng},
  booktitle={International Conference on Learning Representations},
  volume={2024},
  pages={54158--54183},
  year={2024}
}

@inproceedings{opera,
  title={Opera: Alleviating hallucination in multi-modal large language models via over-trust penalty and retrospection-allocation},
  author={Huang, Qidong and Dong, Xiaoyi and Zhang, Pan and Wang, Bin and He, Conghui and Wang, Jiaqi and Lin, Dahua and Zhang, Weiming and Yu, Nenghai},
  booktitle={Proceedings of the IEEE/CVF Conference on Computer Vision and Pattern Recognition},
  pages={13418--13427},
  year={2024}
}

@inproceedings{dimmm,
  title={Hallucination-aware multimodal benchmark for gastrointestinal image analysis with large vision-language models},
  author={Khanal, Bidur and Pokhrel, Sandesh and Bhandari, Sanjay and Rana, Ramesh and Shrestha, Nikesh and Gurung, Ram B and Linte, Cristian and Watson, Angus and Shrestha, Yash R and Bhattarai, Binod},
  booktitle={International Conference on Medical Image Computing and Computer-Assisted Intervention},
  pages={235--245},
  year={2025},
  organization={Springer}
}

@inproceedings{detr,
  title={End-to-end object detection with transformers},
  author={Carion, Nicolas and Massa, Francisco and Synnaeve, Gabriel and Usunier, Nicolas and Kirillov, Alexander and Zagoruyko, Sergey},
  booktitle={European conference on computer vision},
  pages={213--229},
  year={2020},
  organization={Springer}
}

@inproceedings{slake,
  title={Slake: A semantically-labeled knowledge-enhanced dataset for medical visual question answering},
  author={Liu, Bo and Zhan, Li-Ming and Xu, Li and Ma, Lin and Yang, Yan and Wu, Xiao-Ming},
  booktitle={2021 IEEE 18th international symposium on biomedical imaging (ISBI)},
  pages={1650--1654},
  year={2021},
  organization={IEEE}
}

@article{mimic,
  title={MIMIC-CXR, a de-identified publicly available database of chest radiographs with free-text reports},
  author={Johnson, Alistair EW and Pollard, Tom J and Berkowitz, Seth J and Greenbaum, Nathaniel R and Lungren, Matthew P and Deng, Chih-ying and Mark, Roger G and Horng, Steven},
  journal={Scientific data},
  volume={6},
  number={1},
  pages={317},
  year={2019},
  publisher={Nature Publishing Group UK London}
}

@article{medsam3,
  title={MedSAM3: Delving into Segment Anything with Medical Concepts},
  author={Liu, Anglin and Xue, Rundong and Cao, Xu R and Shen, Yifan and Lu, Yi and Li, Xiang and Chen, Qianqian and Chen, Jintai},
  journal={arXiv preprint arXiv:2511.19046},
  year={2025}
}

@inproceedings{huatuo,
  title={Towards injecting medical visual knowledge into multimodal llms at scale},
  author={Chen, Junying and Gui, Chi and Ouyang, Ruyi and Gao, Anningzhe and Chen, Shunian and Chen, Guiming Hardy and Wang, Xidong and Cai, Zhenyang and Ji, Ke and Wan, Xiang and others},
  booktitle={Proceedings of the 2024 conference on empirical methods in natural language processing},
  pages={7346--7370},
  year={2024}
}

@article{lmed,
  title={Llava-med: Training a large language-and-vision assistant for biomedicine in one day},
  author={Li, Chunyuan and Wong, Cliff and Zhang, Sheng and Usuyama, Naoto and Liu, Haotian and Yang, Jianwei and Naumann, Tristan and Poon, Hoifung and Gao, Jianfeng},
  journal={Advances in Neural Information Processing Systems},
  volume={36},
  pages={28541--28564},
  year={2023}
}

@article{svv,
  title={Vision-language models in diagnostic imaging: review of technical advances, clinical validation, and practical deployment},
  author={Dutta, Niharika and Bose, Kartik and Syailendra, Emir and Chu, Linda and Gupta, Pankaj},
  journal={International Journal of Medical Informatics},
  pages={106227},
  year={2025},
  publisher={Elsevier}
}

@inproceedings{utt,
  title={Uncertainty-driven expert control: Enhancing the reliability of medical vision-language models},
  author={Liang, Xiao and Wang, Di and Jiao, Zhicheng and Li, Ronghan and Yang, Pengfei and Wang, Quan and Chua, Tat-Seng},
  booktitle={Proceedings of the IEEE/CVF International Conference on Computer Vision},
  pages={21144--21154},
  year={2025}
}

@inproceedings{bibmm,
  title={Trustworthy Disentangled Framework for Multi-Label Medical Image Classification with Multimodal Refinement},
  author={Chen, Yingyu and Yang, Ziyuan and Huang, Yongqiang and Yang, Xulei and Yeo, Siyong and Zhang, Yi},
  booktitle={2025 IEEE International Conference on Bioinformatics and Biomedicine (BIBM)},
  pages={839--846},
  year={2025},
  organization={IEEE}
}

@inproceedings{mk2,
  title={Trustworthy few-shot transfer of medical VLMs through split conformal prediction},
  author={Silva-Rodr{\'\i}guez, Julio and Ben Ayed, Ismail and Dolz, Jose},
  booktitle={International Conference on Medical Image Computing and Computer-Assisted Intervention},
  pages={658--668},
  year={2025},
  organization={Springer}
}

@inproceedings{icll,
  title={Visual description grounding reduces hallucinations and boosts reasoning in lvlms},
  author={Ghosh, Sreyan and Evuru, Chandra Kiran and Kumar, Sonal and Tyagi, Utkarsh and Nieto, Oriol and Jin, Zeyu and Manocha, Dinesh},
  booktitle={International Conference on Learning Representations},
  volume={2025},
  pages={66510--66547},
  year={2025}
}

@inproceedings{ttr,
  title={Multi-modal hallucination control by visual information grounding},
  author={Favero, Alessandro and Zancato, Luca and Trager, Matthew and Choudhary, Siddharth and Perera, Pramuditha and Achille, Alessandro and Swaminathan, Ashwin and Soatto, Stefano},
  booktitle={Proceedings of the IEEE/CVF Conference on Computer Vision and Pattern Recognition},
  pages={14303--14312},
  year={2024}
}

@inproceedings{npp,
  title={Specvlm: Enhancing speculative decoding of video llms via verifier-guided token pruning},
  author={Ji, Yicheng and Zhang, Jun and Xia, Heming and Chen, Jinpeng and Shou, Lidan and Chen, Gang and Li, Huan},
  booktitle={Proceedings of the 2025 Conference on Empirical Methods in Natural Language Processing},
  pages={7216--7230},
  year={2025}
}

@article{skz,
  title={Accelerating non-maximum suppression: a graph theory perspective},
  author={Si, King-Siong and Sun, Lu and Zhang, Weizhan and Gong, Tieliang and Wang, Jiahao and Liu, Jiang and Sun, Hao},
  journal={Advances in Neural Information Processing Systems},
  volume={37},
  pages={121992--122028},
  year={2024}
}

@inproceedings{mmc,
  title={Chexpo: Preference optimization for chest x-ray vlms with counterfactual rationale},
  author={Liang, Xiao and Hu, Jiawei and Wang, Di and Ma, Zhi and Zhao, Lin and Li, Ronghan and Wan, Bo and Wang, Quan},
  booktitle={Proceedings of the 33rd ACM International Conference on Multimedia},
  pages={2606--2615},
  year={2025}
}

@inproceedings{allin,
  title={Med-VLM: Enhancing Medical Image Segmentation Accuracy through Vision-Language Model},
  author={Zhao, Yihao and Zhong, Enhao and Yuan, Cuiyun and Li, Yang and Zhao, Man and Li, Chunxia and Hu, Jun and Liu, Wei and Liu, Chenbin},
  booktitle={Proceedings of the IEEE/CVF International Conference on Computer Vision},
  pages={7283--7293},
  year={2025}
}

@inproceedings{time,
  title={Vila-m3: Enhancing vision-language models with medical expert knowledge},
  author={Nath, Vishwesh and Li, Wenqi and Yang, Dong and Myronenko, Andriy and Zheng, Mingxin and Lu, Yao and Liu, Zhijian and Yin, Hongxu and Law, Yee Man and Tang, Yucheng and others},
  booktitle={Proceedings of the Computer Vision and Pattern Recognition Conference},
  pages={14788--14798},
  year={2025}
}

@article{kang2026vispec,
  title={Vispec: Accelerating vision-language models with vision-aware speculative decoding},
  author={Kang, Jialiang and Shu, Han and Li, Wenshuo and Zhai, Yingjie and Chen, Xinghao},
  journal={Advances in Neural Information Processing Systems},
  volume={38},
  pages={115511--115532},
  year={2026}
}

\end{document}